\documentclass[runningheads]{llncs}
\usepackage[T1]{fontenc}
\usepackage{graphicx}
\usepackage{microtype} 
\usepackage{booktabs}  
\usepackage{amsmath}
\usepackage{algorithm}
\usepackage{algpseudocode}
\usepackage[usenames,dvipsnames]{color}
\usepackage[toc]{appendix}
\usepackage{soul}

\DeclareMathOperator*{\argmax}{arg\,max}

\algnewcommand{\IIf}[1]{\State\algorithmicif\ #1\ \algorithmicthen}
\algnewcommand{\EndIIf}{\unskip\ \algorithmicend\ \algorithmicif}

\begin{document}
\title{A Meta-Level Learning Algorithm for Sequential Hyper-Parameter Space Reduction in AutoML}
\titlerunning{A Meta-Level Learning Algorithm for Hyper-Parameter Space Reduction}
%
\author{
Giorgos Borboudakis\inst{1} \and
Paulos Charonyktakis\inst{1} \and
Konstantinos Paraschakis\inst{1} \and
Ioannis Tsamardinos\inst{1,2}
}

\authorrunning{G. Borboudakis et al.}
%
\institute{JADBio - Gnosis DA S.A. \and
Computer Science Department - University of Crete
\email{\{borbudak,haronykt,kparaschakis,tsamard\}@jadbio.com}}
\maketitle              
\begin{abstract}
AutoML platforms have numerous options for the algorithms to try for each step of the analysis, i.e., different possible algorithms for imputation, transformations, feature selection, and modelling.
Finding the optimal combination of algorithms and hyper-parameter values is computationally expensive, as the number of combinations to explore leads to an exponential explosion of the space.
In this paper, we present the Sequential Hyper-parameter Space Reduction (SHSR) algorithm that reduces the space for an AutoML tool with negligible drop in its predictive performance.
SHSR is a meta-level learning algorithm that analyzes past runs of an AutoML tool on several datasets and learns which hyper-parameter values to filter out from consideration on a new dataset to analyze. SHSR is evaluated on 284 classification and 375 regression problems, showing an approximate $30\%$ reduction in execution time with a performance drop of less than $0.1\%$.

\end{abstract}

\section{Introduction}

AutoML platforms for predictive modelling try to solve the Combined Algorithm Selection and Hyper-parameter Optimization (CASH) \cite{Thornton2013} problem. CASH optimizes the algorithmic choices, as well as the hyper-parameter values of the machine learning pipeline (hereafter called a \textbf{configuration}) that produces the final model. Each configuration may contain several steps, such as algorithms for missing value imputation, feature transformation, feature extraction, feature selection, and of course, predictive modelling. All such choices can be represented with hyper-parameters which form the decision variables of an optimization problem with objective function the out-of-sample predictive performance of the final model, called Hyper-Parameter Optimization (HPO). The hyper-parameters form the configuration space over which to optimize. 

There have been at least two complementary approaches to address this optimization. The first approach is to employ black-box optimization strategies to search the configuration space, such as grid search, random search, and Sequential Bayesian Optimization \cite{Snoek2012}. These algorithms evaluate the configurations on the dataset to analyze by training models. The second approach is try to reduce or prioritize the exploration of the configuration space based on prior information. Such algorithms, called meta-level learning algorithms, analyze the past performance of configurations on other datasets and try to predict their performance on a new dataset based on its characteristics, called meta-features \cite{Rivolli2022}. Obviously, the two approaches can be combined with the meta-level algorithm selecting a subset of configurations to explore, and the HPO algorithm performing an intelligent search within this reduced space. 

When optimizing over a continuous hyper-parameter, e.g., the cost $C$ of a Support Vector Machine (SVM), several reasonable assumptions to facilitate to explore the configuration space have been proposed, e.g., that neighboring $C$ values will lead to correlated model performances. In the Bayesian Sequential Optimization these assumptions are captured by the kernel function used in the Gaussian Process that fits the performance landscape. However, for discrete choices and hyper-parameters, particularly the choice of which algorithm to use at each analysis step, it is less clear if and which assumptions are reasonable. Will an SVM perform well on a specific dataset, given the performance of the Decision Tree? Meta-level learning algorithms can be proven invaluable when optimizing categorical hyper-parameters.

In this paper, we propose such an algorithm called \textbf{Sequential Hyper-Parameter Space Reduction} (SHSR). 
SHSR is a meta-level learning algorithm that analyzes the past performances and execution times of configurations on a corpus of datasets, stored in matrices $\mathbf{P}$ and $\mathbf{E}$, respectively.
It learns which configurations can be safely removed from consideration given the meta-features of the dataset, without affecting the predictive performance of the final model (within some user-defined tolerance threshold), while simultaneously trying to minimize execution time.
It then recursively applies this step to the remaining configurations.
By removing such values, SHSR exponentially reduces the discrete part of the configuration space. 
To apply SHSR, it is required that a set of configurations is run on a corpus of datasets, and their performances and execution times are measured and stored in matrices $\mathbf{P}$ and $\mathbf{E}$, respectively. 

SHSR is evaluated on 284 classification and 375 regression problems on real public datasets. The trade-offs between the tolerance threshold vs. the computational savings are explored. In addition, it is shown that the algorithm performs well when provided with incomplete matrices $\mathbf{P}$ and $\mathbf{E}$, i.e., when configurations are run only on a small fraction of the datasets, practically providing the same results even if only 20\% of the values in $\mathbf{P}$ and $\mathbf{E}$ are present. For a very strict tolerance threshold, SHSR achieves a relative drop in predictive performance of less than 0.1\%, with 30\% computational savings. We note that \textit{time savings are measured with respect to a simple grid search, where all choices are discrete}. For less strict thresholds, SHSR achieves computational savings of 50\% and 40\% with a relative performance drop of 1.5\% and 0.1\%) for classification and regression, respectively. 

This paper is structured as follows: Section 2 overviews the literature of meta-level learning and dataset characterization. Section 3 presents our proposed algorithm, SHSR, while Section 4 describes SHSR's experimental evaluation. Specifically, in Section 4.1, we describe the experimental setup, while section 4.2 presents the evaluation results. Finally, Section 5 summarizes our conclusions and future work, while Section 6 discusses the limitations of this work.

\section{Related Work}

Meta-level learning, often called meta-learning, is a heavily overloaded term, and has been used to refer to different problems in the literature, such as algorithm recommendation, ensembling and transfer learning \cite{Lemke2015}. 
We will hereafter focus on the \textbf{Algorithm Recommendation Problem} \cite{Furnkranz2002}, which deals with learning to recommend configurations to try for a new problem, based on past runs. 
This is typically done by first characterizing datasets using measures (also called \textbf{meta-features}) that describe their properties, and then learning to predict the predictive performance and/or running time of configurations.
Next, we will provide a brief overview of different meta-features and meta-level learning methods.
Interested readers may refer to \cite{Vanschoren2018} for an overview of meta-level learning, and to \cite{Kedziora2020} on how it is used in the context of AutoML.

\subsection{Dataset Characterization}

\cite{Rendell1987,Rendell1989} introduced the idea of dataset characterization for meta-level learning, and studied the effect of simple measures on the performance of learning algorithms.
\cite{Michie1994} describe several dataset characterization measures, and group them into simple (e.g., number of samples), statistical (e.g., average absolute correlation between feature pairs) and information-theoretic measures (e.g., average entropy of features).
Additional categories and types of measures have been introduced over time: model-based measures, which are extracted from models induced from the meta-features (e.g., the tree depth of a decision tree trained on the meta-features), landmarking measures, which use the performance of simple and fast learning algorithms (e.g., accuracy of Naive Bayes), and other measures (e.g., silhouette index of k-means clustering).
We refer the reader to \cite{Rivolli2022} for a comprehensive review of meta-features.

\subsection{Meta-level Learning}

To the best of our knowledge, the algorithm recommendation problem was first addressed in \cite{Rendell1987}. The authors used simple meta-features, and studied whether learning to recommend algorithms based on them leads to improvements.
\cite{Aha1992} extended that work, by (a) considering additional simple meta-features and (b) introducing the idea of learning interpretable rules for algorithm recommendation, and evaluated the method on synthetic data.
\cite{Brazdil1994} were the first to apply meta-level learning on real-world data. They used simple, statistical and information-theoretic meta-features, and applied decision trees to learn whether an algorithm should be applied on a given dataset or not.
The work was subsequently extended to use regression and instance-based models to directly predict the predictive performance of algorithms on new datasets \cite{Gama1995}. \cite{Brazdil2003} use a KNN based algorithm on dataset meta-features to find similarities and then they apply a performance/computational time multi-criterion to rank algorithms (more on this in our Comparison Section below).
More recently, collaborative filtering methods  \cite{misir2017,hosoya2018,Yang2018,Yang2020} were proposed to solve the algorithm recommendation problem, inspired by the Netflix challenge.
These methods allow for incomplete inputs (i.e., not all configurations are run on all datasets).
However, they suffer from the cold-start problem, but there exist approaches that try to avoid that by using meta-features \cite{misir2017}.
For a detailed review of meta-level learning methods, we point the reader to \cite{Luo2016,Khan2020}.

\section{The Sequential Hyper-parameter Space Reduction Algorithm for Meta-level Learning}

In this section we introduce the \textbf{Sequential Hyper-parameter Space Reduction} algorithm (SHSR) for algorithm recommendation.
SHSR uses the predictive performances and execution times of groups of configurations on past analyses, and returns models which are used to filter out unpromising groups, while simultaneously trying to minimize execution time.
Configuration groups can contain one or multiple configurations (e.g., all configurations with the same modelling algorithm), are not necessarily mutually exclusive (i.e., the same configuration might be present in multiple groups), and the input can be partial (i.e., results for some configurations might not be present for all past analyses).
SHSR is shown in Algorithm~\ref{alg:shsr}.
We proceed with a detailed description of SHSR.

Let $G$, $D$, and $F$ denote the number of configuration groups, datasets, and meta-features, respectively.
SHSR takes as input: (a) $G \times D$ matrices $\mathbf{P}$ and $\mathbf{E}$ containing the performance ratios and execution times of all configuration groups, (b) an $F \times D$ matrix $\mathbf{X}$ of meta-features, (c) a threshold $T$, and (d) a list of $G$ sets $\textproc{Active}$, with $\textproc{Active}[g]$ initialized to all datasets for which results for $g$ are present.
For a given group $g$ and dataset $d$, the performance ratio $\mathbf{P}_{g,d}$ is defined as the maximum performance over configurations in group $g$ on dataset $d$, divided by the maximum performance over all configurations.
The execution time $\mathbf{P}_{g,d}$ is defined as the sum of execution times of all configurations in $g$.
The output is a sequence of models $(\textproc{Model}[g_1], \dots, \textproc{Model}[g_k])$, where $\textproc{Model}[g]$ is a model predicting the performance ratio achieved without group $g$ on a dataset $d$ based on its meta-features $\mathbf{X}_{*,d}$.

SHSR starts by creating one model per configuration group $g$, and computing the time saved if that group were to be removed from consideration.
For this, an outcome $\mathbf{y}$ is created, which contains the maximum performance for each dataset in $\textproc{Active}[g]$ over all groups except $g$, and a model $\textproc{Model}[g]$ for $\mathbf{y}$ is fitted using meta-features $\mathbf{X}$.
Next, $\textproc{Covered}[g]$ is computed, by applying $\textproc{Model}[g]$ on all $\textproc{Active}[g]$, and selecting only the ones for which the model predicts a performance ratio at least as large as $T$.
In other words, $\textproc{Covered}[g]$ contains all datasets for which group $g$ can be excluded, as the remaining groups are sufficient to achieve a high performance ratio.
Finally, the time saved $\textproc{TimeSavings}[g]$ is computed as the sum of execution times of all $\textproc{Covered}[g]$.
Once this is done, the group $g^*$ with the highest time savings is selected, and SHSR is called recursively with $\textproc{Covered}[g^*]$ removed from $\textproc{Active}[g]$ (in the algorithm we slightly abuse notation for the sake of brevity).
SHSR stops once no more time savings can be achieved, which happens if no more datasets can be covered by the removal of any group.

Finally, to decide which groups to run on a new dataset $d'$, the models $(\textproc{Model}[g_1], \dots, \textproc{Model}[g_k])$ are applied in sequence, and a group $g_i$ is removed from consideration if $\textproc{Predict}(\textproc{Model}[g_i], \mathbf{X_{*, d'}}) \geq T$.

\begin{algorithm}[!t]
\caption{Sequential Hyper-parameter Space Reduction}\label{alg:shsr}
\textbf{Constants:} Performance ratios $\mathbf{P}$, Execution times $\mathbf{E}$, Meta-features $\mathbf{X}$, 
Threshold $T$ 
\\
\textbf{Input:} Active datasets per configuration group $\textproc{Active}$
\\
\textbf{Output:} Sequence of regression models
\begin{algorithmic}[1]
\For {each configuration group $g$}
    \State $\mathbf{y}_d \gets \max_{i} \mathbf{P}_{i, d}, i \neq g, \forall d \in \textproc{Active}[g]$
    \State $\textproc{Model}[g] \gets \textproc{FitModel}(\mathbf{y}, \mathbf{X})$ \label{shsr:line:model}
    \State $\textproc{Covered}[g] \gets \{d \mid d \in \textproc{Active}[g] \wedge \textproc{Predict}(\textproc{Model}[g], \mathbf{X_{*, d}}) \geq T \}$
    \State $\textproc{TimeSavings}[g] \gets \sum_d \mathbf{E}_{g, d}, d \in \textproc{Covered}[g]$
\EndFor
\State $g^* \gets \argmax_{g} \textproc{TimeSavings}[g]$
\If{$\textproc{TimeSavings}[g^*] = 0$}
    \State \Return $\emptyset$
\EndIf
\State \Return $(\textproc{Model}[g^*]) \cup \textproc{SHSR}(\textproc{Active} \setminus \textproc{Covered}[g^*])$
\end{algorithmic}
\end{algorithm}

\section{Experimental Evaluation}

In this section we evaluate the ability of SHSR to filter out groups of configurations, and measure its impact on running time and predictive performance. 
To this end, we collected classification and regression datasets and trained a set of predefined configurations on them, in order to obtain performance estimates and execution times for SHSR.
The full analysis required training $\sim$85M models, taking a total of $\sim$58K core hours.
We proceed with a detailed description of the experimental setup. Results are presented in Section~\ref{sec:results}.

\subsection{Experimental Setup}
\label{sec:exp_setup}

\subsubsection{Datasets}

\begin{figure}[t!]
\centering
  \includegraphics[scale=0.375]{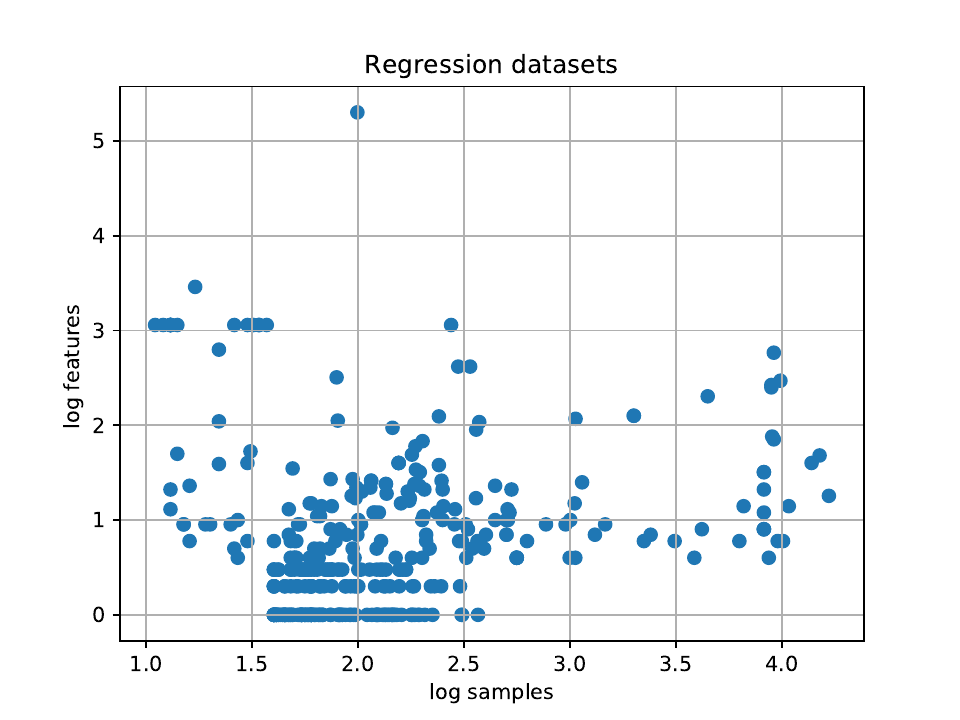}%
  \includegraphics[scale=0.375]{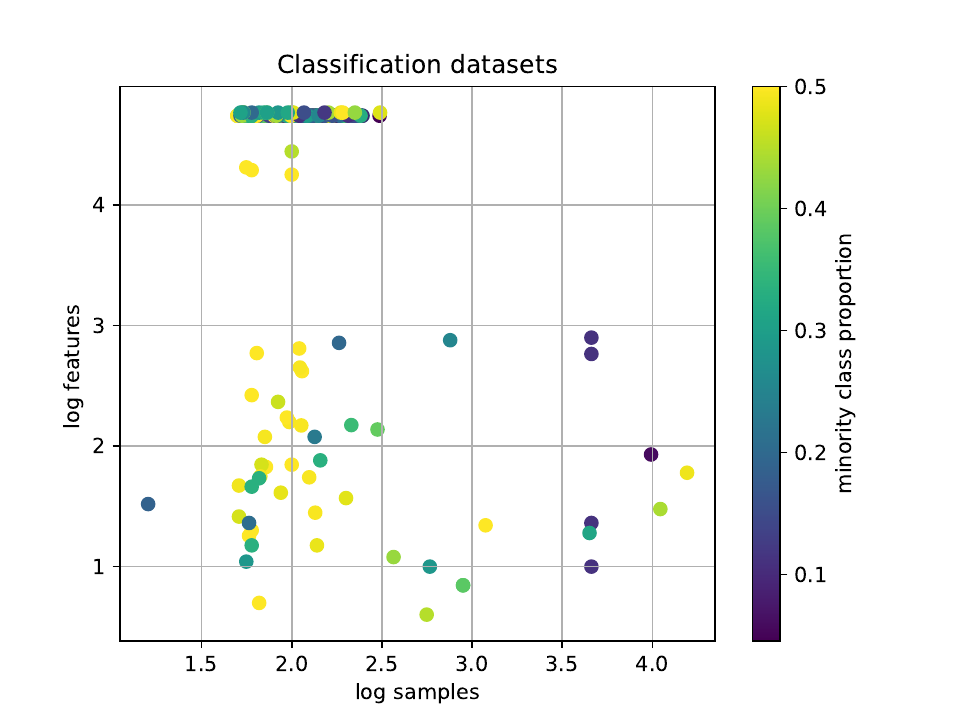}%
  \caption{Sample size vs feature size for regression and classification datasets. The x-axis shows the sample size, while the y-axis shows the feature size. For the classification datasets, the color intensity varies depending on the class distribution. Both axes are in $\log_{10}$ scale.}
  \label{fig:datasets}
\end{figure}

We collected a total of 659 datasets, out of which 284 are binary classification problems and 375 are regression problems.
Datasets were selected to cover a wide range of problems, with varying sample sizes, number of variables and, in case of classification problems, class distributions; see Fig.~\ref{fig:datasets} for a summary.
All datasets were downloaded from OpenML \cite{OpenML2013} (licensed under CC-BY 4.0) and from BioDataome \cite{Lakiotaki2018} (publicly available).
A list of all datasets and their sources can be found on the project's GitHub repository\footnote{https://github.com/JADBio/SHSR}.

\subsubsection{Implementation}

For model training we used JADBio \cite{Tsamardinos2022}. 
JADBio is an automated machine learning platform specifically designed for life scientists and molecular data. 
It uses a fully automated machine learning protocol to select the best combination of preprocessing, feature selection and modelling algorithms, and returns a predictive model and an estimate of its predictive performance.
Any features that could interfere with results (e.g., early dropping of unpromising configurations \cite{Tsamardinos2018}) were disabled.
Also, in order to get accurate timing results, only a single CPU core was used per configuration, and analyses were not run in parallel.

Everything else was implemented in Python and is available on GitHub.
For machine model training, we used the scikit-learn package \cite{scikit-learn}.
This includes code for the implementation of SHSR, as well as code for producing all results.

\subsubsection{Algorithms and Configurations}

For preprocessing, JADBio uses standardization and mean imputation for continuous variables and mode imputation for the categorical ones.
As feature selection algorithms, the Statistical Equivalent Signatures (SES) algorithm \cite{Lagani2017} and Lasso \cite{Tibshirani1996} were employed. 
Finally, for modelling JADBio uses $L_2$-regularized linear and logistic regression \cite{Hoerl1970}, Support Vector Machines (SVM) \cite{Boser1992} and Support Vector Regression (SVR) with Linear, Polynomial, and Gaussian kernels, Decision Trees \cite{Breiman1984}, and Random Forests \cite{Breiman2001}.

We used the default grid search parameters of JADBio with the tuning parameter set to ``extensive'': 6 hyper-parameter combinations for SES, 7 for Lasso, 7 for linear and logistic regression, 150 for SVMs, 1500 for SVRs, 15 for Decision Trees and 60 for Random Forests\footnote{The complete list of hyper-parameters is available on GitHub.}.
Configurations were obtained by taking all combinations of algorithms and hyper-parameter values, and constraining them to contain one or multiple transformations (imputation is always included, if applicable, and takes precedence over standardization), one feature selection algorithm, and one modelling algorithm.
The total number of configurations were 2983 and 8633 for classification and regression tasks, respectively.

Based on the above configurations, we created a total of 21 configuration groups: (a) one group per feature selection algorithm (13 in total, one per hyper-parameter combination), and (b) one group per modelling algorithm (8 in total; SVMs with different kernels and polynomial kernels with different degrees are considered separate algorithms).
Each such group contains only the subset of configurations for which the respective algorithm was used (e.g., the random forest group contains all results with random forest, irrespective of the feature selection algorithm used).

For performance estimation, we ran JADBio with 10-fold cross-validation, repeated at most 20 times (a stopping criterion is applied when a plateau is reached during the repetitions) \cite{Tsamardinos2018}.
We note that JADBio returns unbiased (and typically conservative) performance estimates, by applying the BBC-CV algorithm \cite{Tsamardinos2018} on the out-of-sample predictions from cross-validation, without the need of using a test set or expensive techniques like nested cross-validation.

At this point we note that the execution time of feature selection algorithms was only counted once per fold and not per configuration, as in practice feature selection has to be performed only once for a given dataset and hyper-parameter combination, regardless of the number of subsequently trained models.

\subsubsection{Evaluation of SHSR}

\begin{table}[t!]
\caption{Meta-features used in the experiments.}
\centering
\scriptsize
\begin{tabular}{ |l|l| }
\hline
 Meta-feature & Description \\ 
 \hline
 n\_samples & Number of samples \\
 n\_features & Number of features \\
 samples\_to\_features & Number of samples to number of features ratio \\
 total\_missing & Number of missing values \\
 total\_missing\_f & Proportion of missing values \\
 samples\_with\_any\_missing & Number of samples with at least one missing value \\
 samples\_with\_any\_missing\_f & Proportion of samples with at least one missing value \\
 categorical\_features & Number of categorical features \\
 numerical\_features & Number of numerical features \\
 target\_majority\_class\_instances & Number of samples for majority class \\
 target\_majority\_class\_f & Proportion of samples for majority class \\
 target\_minority\_class\_instances & Number of samples for minority class \\
 target\_minority\_class\_f & Proportion of samples for minority class \\
 categorical\_to\_numerical & Ratio of categorical to numerical features \\
 silhouette\_k & Silhouette index of k-means clustering \\
 pca\_p & Number of components that explain $p\%$ of variance \\ 
 \hline
\end{tabular}
\label{tbl:meta_features}
\end{table}

Table~\ref{tbl:meta_features} lists the meta-features used for the analysis \cite{Rivolli2022}. 
We used 14 simple measures, along with the silhouette index of k-means clustering for $k \in \{2,3,\dots,10\}$ and the number of PCA components required to explain $p\%$ of the variance of a dataset for $p \in \{60, 70, 80, 90\}$, resulting in a total of 27 meta-features.
We note that, as the types of meta-features used are not part of SHSR, we mainly chose simple measures because they are easy and fast to compute.

To evaluate SHSR, we randomly keep out 10\% of the datasets for testing, and executed SHSR on the remaining 90\%. 
For each held-out dataset, we applied the model returned by SHSR to recommend a set of configurations to compute the predictive performances and running times.
The procedure was repeated 20 times and averages are reported.

Finally, as a regression model in SHSR (see Line~\ref{shsr:line:model}) we used regression trees, and tuned them using 5-fold cross-validation, by optimizing over min\_samples\_leaf $\in \{3, 5, 7\}$, as well as over the solution path returned by the minimal cost-complexity pruning algorithm \cite{scikit-learn}.
The reason we chose decision trees is because they are interpretable, non-linear and have a low computational cost.

\subsection{Results}
\label{sec:results}

We ran experiments to investigate the effect of the threshold $T$ on predictive performance and time savings when using SHSR to select configurations.
Next, we evaluate how SHSR performs when only a random subset of the results is available.
Finally, we compare SHSR to a trivial algorithm which randomly removes a proportion of configurations.
Results and additional details about the experiments are presented below.

\subsubsection{Effect of Threshold $T$ on Predictive Performance and Execution Time}

\begin{figure}[t!]
\centering
  \includegraphics[scale=0.375]{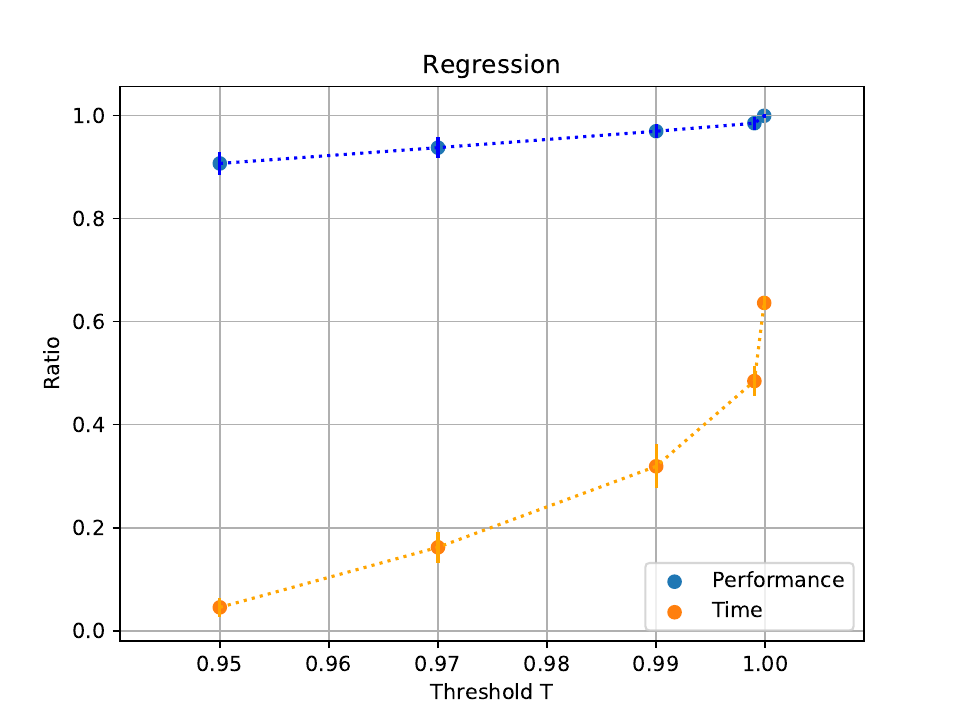}%
  \includegraphics[scale=0.375]{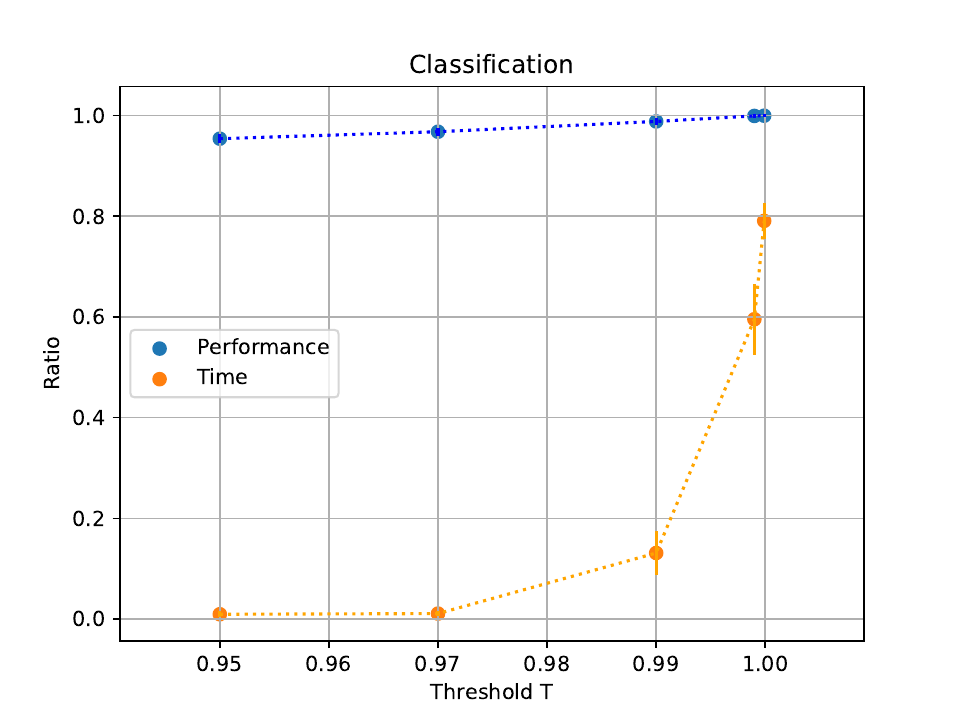}%
  \caption{Effect of threshold $T$ on predictive performance and execution time. The x-axis shows the threshold, and the y-axis shows the ratio between the predictive performance (execution time) using the configurations returned by SHSR, relative to using all configurations. Error bars show the 95\% Gaussian confidence intervals for the mean, resulting from 20 runs of the experiment. We observe that SHSR leads to a significant reduction in execution time, with minimal drop in predictive performance.}
  \label{fig:results_threshold}
\end{figure}

We varied the threshold hyper-parameter $T$ of SHSR to investigate the trade-off between predictive performance and running time. 
Specifically, we considered thresholds in $\{0.95, 0.97, 0.99, 0.999, 0.9999\}$.
Lower values were not tried, as they would lead to a significant drop in performance, with minimal time savings.

The results are summarized in Fig.~\ref{fig:results_threshold}.
As one would expect, with increasing $T$, performance increases while execution time decreases.
Furthermore, we observe that for very low values of $T$ there is a negligible loss in performance, while still providing significant time savings.
For instance, a threshold of $0.999$ leads to an average performance loss of $\sim1.5\%$ for regression and $\sim0.1$ for classification problems, while saving $\sim50\%$ and $\sim40\%$ time respectively.
On the other hand, if one was mainly interested in execution time, they could use a lower threshold, sacrificing predictive performance for a reduction in execution time.
For example, a threshold of $0.95$ reduces time by $\sim95\%$ and $\sim99\%$ for regression and classification tasks respectively, while retaining $\sim91\%$ and $\sim95\%$ of the predictive performance.

\subsubsection{Evaluation on Partial Results}

\begin{figure}[t!]
\centering
  \includegraphics[scale=0.375]{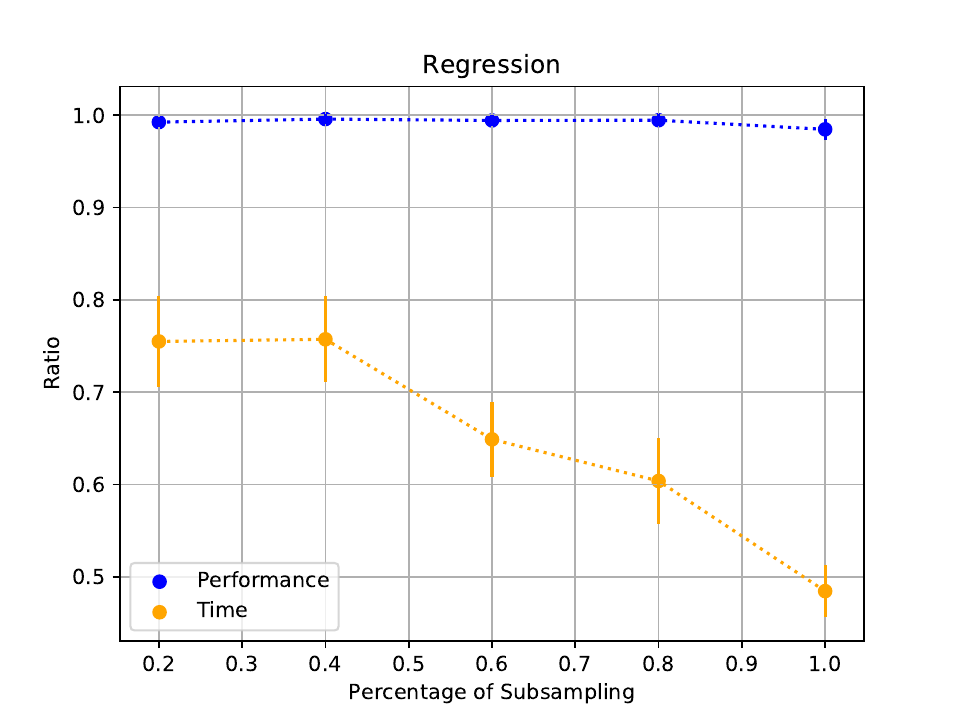}%
  \includegraphics[scale=0.375]{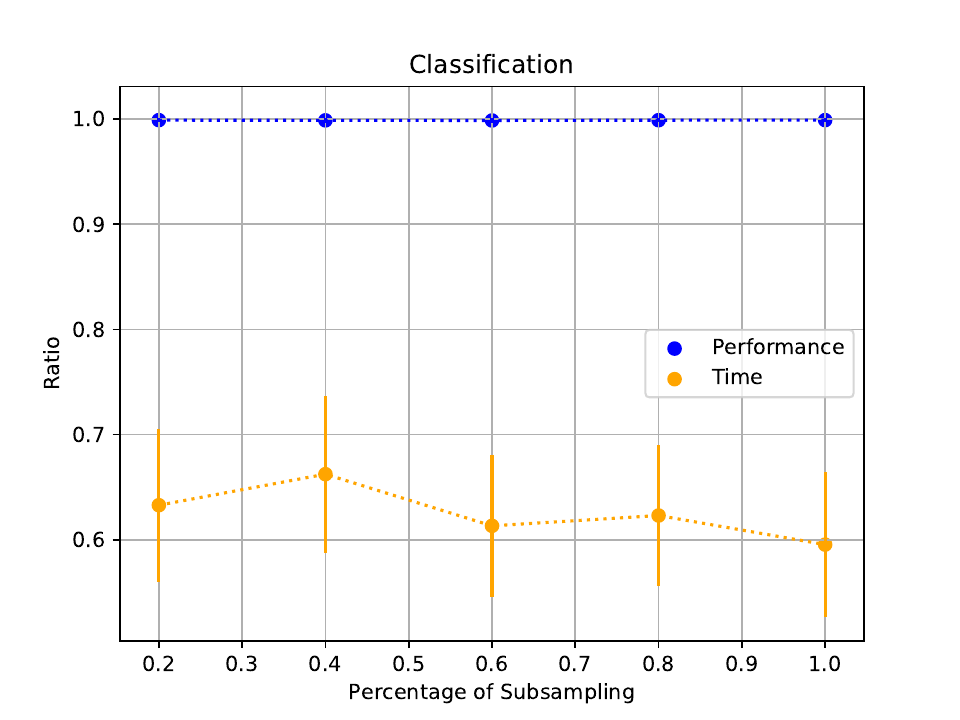}%
  \caption{Effect of running SHSR with partial results on predictive performance and execution time. The x-axis shows the proportion of used results, and the y-axis shows the ratio between the predictive performance (execution time) using the configurations returned by SHSR, relative to using all configurations. Error bars show the 95\% Gaussian confidence intervals for the mean, resulting from 20 runs of the experiment. We observe that SHSR is able to perform well even with partial results, and that it performs better the more results are available, as expected.}
  \label{fig:results_subsampling}
\end{figure}

In this experiment we evaluate SHSR when run with partial results (i.e., when only a subset of configurations have been applied on each dataset).
We simulated that scenario by randomly sub-sampling $\{20\%, 40\%, 60\%, 80\%\}$ of all results.
Note that sub-sampling was performed on all results, not on each dataset or configuration.
The threshold $T$ was set to $0.999$ for this experiment. 

Fig.~\ref{fig:results_subsampling} shows the results.
We can see that SHSR performs well, even if only 20\% of all results are available.
Furthermore, we observe that execution time is decreasing the more results are available to SHSR, especially for regression tasks, where it drops from $\sim75\%$ of the original running time when 20\% of the results are used, to $\sim50\%$ when all results are used.
\textit{This confirms that SHSR is able to make better decisions the more data are available, as one would expect}.

\subsubsection{Comparison against \cite{Brazdil2003} and random elimination of configurations}

As a final experiment, we compared SHSR to the KNN based meta-level algorithm in \cite{Brazdil2003}, as well as a simple baseline which randomly eliminates a percentage of configurations. The authors in \cite{Brazdil2003} apply KNN on the meta-features of the datasets in order to locate the $N$ most similar from a pool of already evaluated datasets for a new incoming dataset. Subsequently they apply a pairwise multi-criterion (performance/time) metric to compare all available configurations on the training datasets and a configuration specific relative score is being calculated. The latter can be used to rank configurations and select the top ranking to run on the new dataset. Regarding the random elimination, we considered randomly removing $\{50\%, 60\%, 70\%, 80\%, 90\%, 95\%, 97\%, 99\%\}$ of configurations.
SHSR was executed with $T$ taking the values $0.95, 0.97, 0.99, 0.999, 0.9999$. The KNN based algorithm was applied with various hyper-parameters (see original paper for a description of them): $N neighbors \in \{1, 3, 10\}, AccD \in \{0.001, 0.01, 0.1\}$. The method was used to rank configurations according to their performance and execution time on test datasets and then the top $m$ ranked configurations were chosen to be applied on the test datasets, where $m \in \{100, 300, 500, 1000, 1500, 2000, 2500\}$ for classification and $m \in \{100, 500, 1000, 2000, 4000, 6000\}$ for regression. Each combination setting was repeated 20 times. 

\begin{figure}[t!]
\centering
  \includegraphics[scale=0.375]{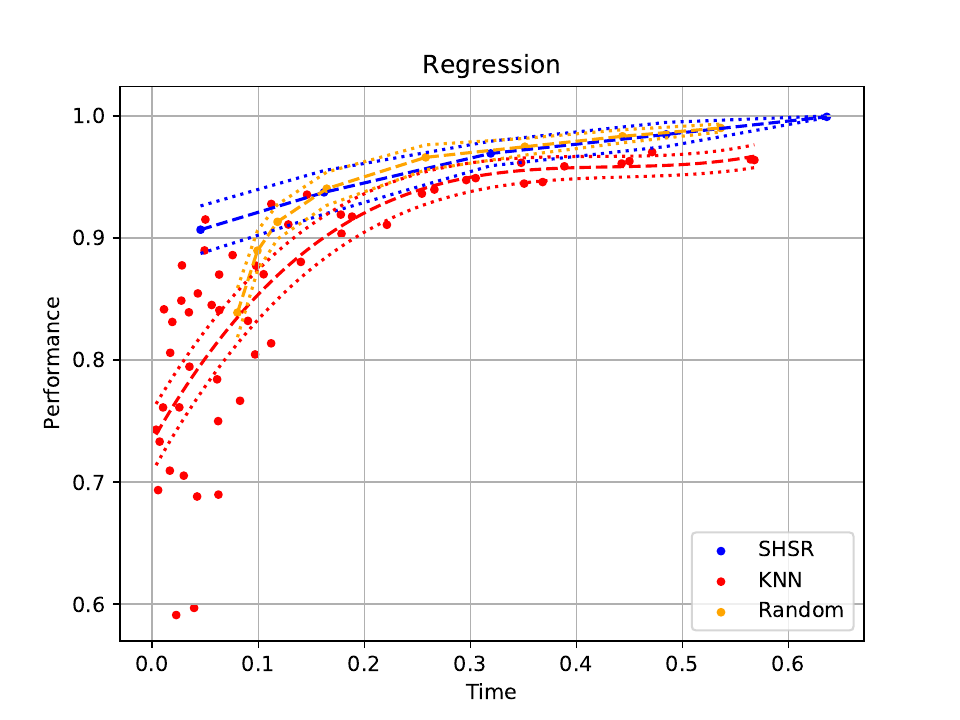}%
  \includegraphics[scale=0.375]{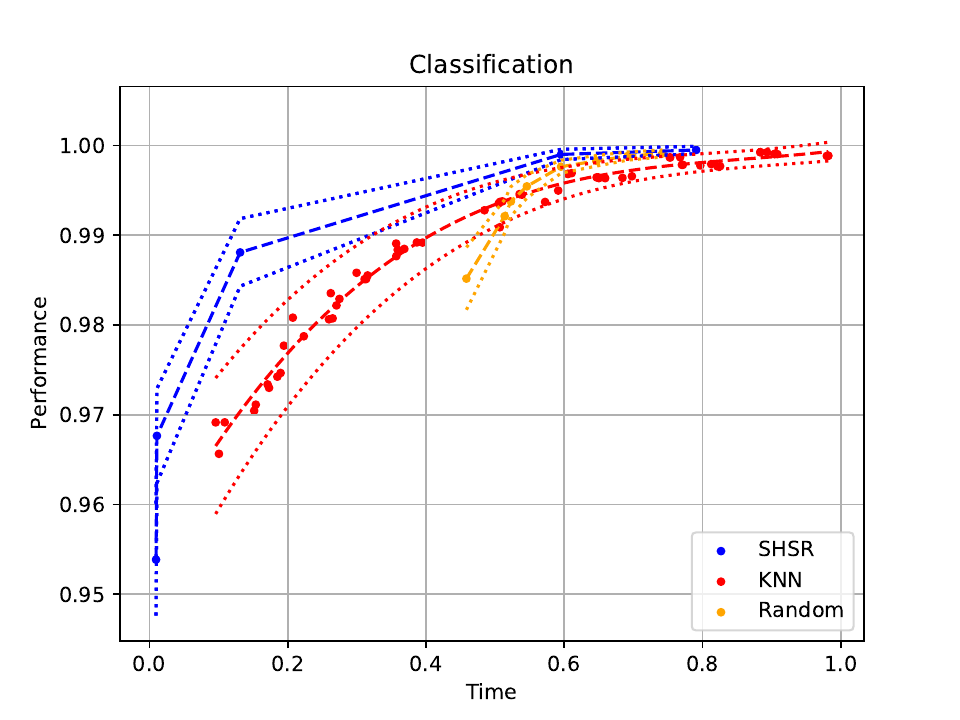}%
  \caption{Comparison between SHSR, KNN based algorithm, and random elimination. The x-axis (y-axis) shows the ratio of execution time (performance) using the configurations returned by the algorithms, relative to using all configurations. Dotted lines show the 95\% Gaussian confidence intervals for the mean, resulting from 20 runs of the experiment. In the case of the KNN based algorithm splines were used to smooth both the mean and the CI lines. All three algorithms perform similarly when few configurations are dropped, with SHSR being superior when many configurations are removed.}
  \label{fig:versus_other}
\end{figure}

The results are shown in Fig.~\ref{fig:versus_other}.
For regression tasks, random elimination of less than $90\%$ of configuration leads to similar results as SHSR, both in terms of predictive performance and running time.
The same holds for classification tasks, when fewer than $80\%$ of configurations are removed.
However, if more configurations are removed, performance of random elimination drops abruptly, with minimal time savings.
We note that the reason running time does not drop linearly with configurations, which would be expected to hold on average using random elimination, is that it is unlikely that a feature selection algorithm instance is completely removed by chance, as this can only happen if all configurations it participates in are removed.
In contrast, SHSR does not suffer from this, as it either keeps an algorithm group or removes it completely.

The KNN based algorithm seems to be performing in-between the other two for classification and slightly worse for regression for less than $90\%$ time reduction. It does exceed random elimination's performance even in regression for heavily reduced configurations sets, though.

We further investigated why random elimination performs that well.
We found that often multiple configurations achieve maximum performance or close to it (i.e., there is a lot of redundancy), making it likely to select a good configuration by chance.
This is explained by the fact that the hyper-parameter configurations used by JADBio are curated and optimized to cover a wide range of problems.
Thus, we would expect random elimination's performance to drop if the proportion of ``lower quality'' configurations increases, while SHSR wouldn't be affected significantly, as it uses the maximum performance of a group to determine its performance.
To better understand that, consider an example with 10000 configurations, 50 out of which are optimal or very close to it, and assume that 9000 of the configurations are randomly removed.
Then, the probability of selecting any optimal configuration by chance is $1 - 0.995^{1000} = 99.33\%$.
If we were to reduce the number of optimal configurations to 25 (i.e., increase the proportion of bad configurations), the probability of selecting any of them drops to $1 - 0.9975^{1000} = 91.82\%$.
Thus, \textit{we argue that in practice SHSR is always preferable over random elimination, especially when used for hyper-parameter configurations that have not been fine-tuned}. SHSR is also clearly superior to the KNN based algorithm in all investigated scenarios.

\section{Discussion, Conclusions, and Future Work}

SHSR introduces a recursive evaluation of choices and their filtering in meta-level learning. It is evaluated on a large cohort of datasets, spanning a wide range of dataset characteristics. It exhibits significant computational savings for a simple grid search, with minimal relative drop in predictive performance. The performance drop can be controlled by a tolerance threshold. The algorithm can be coupled as a filtering step with any HPO strategy when some hyper-parameter domains are discrete.
Learning which values to filter in SHSR is currently evaluated based on a simple Decision Tree algorithm. Any modelling algorithm could be employed instead, of course. However, trees have the advantage of being interpretable. Hence, the decision of the system to drop some algorithmic choices from consideration based on the characteristics of the dataset, can be explained to the user of the AutoML platform. In addition, it can be used to provide intuition into the deficiencies of algorithms in particular situations that may inspire new algorithms. There are numerous future directions to explore, such as coupling SHSR with other HPO algorithms than grid search and employing more powerful models than DTs for filtering.

\section{Limitations}
This study contains some limitations inevitably. First of all, most of the datasets for the classification experiments used in this study are molecular (multi-omics) datasets which are high dimensional and typically contain 
relatively few samples. Secondly, while the algorithm is general and can be applied on any analysis task this study evaluates it only in binary classification and regression tasks. Finally, we evaluated the SHSR employing only Decision Trees for their rule generation feature without testing any other modelling algorithm. 

\subsubsection{Acknowledgements}
The research leading to these results has received funding
from the European Research Council under the European
Union’s Seventh Framework Programme (FP/2007-2013) /
ERC Grant Agreement n. 617393, the Hellenic Foundation for Research and Innovation (H.F.R.I.) under the “First Call for H.F.R.I. Research
Projects to support Faculty members and Researchers and the
procurement of high-cost research equipment grant” (Project
Number: 1941), and the METALASSO project, which is co-financed by the European Union and Greek national funds through the Operational Program Competitiveness, Entrepreneurship and Innovation, under the call RESEARCH–CREATE– INNOVATE (project code: T1EDK-04347).

\bibliographystyle{splncs04}
\bibliography{ref}

\end{document}